\documentclass[journal, 10 pt, twocolumn]{IEEEtran}
\IEEEoverridecommandlockouts  % This command is only needed if you want to use the \thanks command
\usepackage{graphics}             % for pdf, bitmapped graphics files
\usepackage{subfig}
\usepackage{epsfig}                % for postscript graphics files
\usepackage{newtxmath}
\usepackage{times}                % assumes new font selection scheme installed
\usepackage{amsmath}           % assumes amsmath package installed

\usepackage{array}

\usepackage{color}
\usepackage{ntheorem}
\usepackage{bbm}

\usepackage{multirow}

\usepackage{algorithm}
\usepackage[algo2e]{algorithm2e}
\usepackage[noend]{algpseudocode}

\usepackage{lipsum}% just for the example

\makeatletter
\newcommand\fs@betterruled{%
  \def\@fs@cfont{\bfseries}\let\@fs@capt\floatc@ruled
  \def\@fs@pre{\vspace*{5pt}\hrule height.8pt depth0pt \kern2pt}%
  \def\@fs@post{\kern2pt\hrule\relax}%
  \def\@fs@mid{\kern2pt\hrule\kern2pt}%
  \let\@fs@iftopcapt\iftrue}
\floatstyle{betterruled}
\restylefloat{algorithm}
\makeatother

\makeatletter
\def\BState{\State\hskip-\ALG@thistlm}
\makeatother

\newcommand{\argmax}{\arg\!\max}

\title{\LARGE \bf
Human-guided Robot Behavior Learning: A GAN-assisted Preference-based Reinforcement Learning Approach}
\author{Huixin Zhan, Feng Tao, and Yongcan Cao
\thanks{The authors are with the Department of Electrical and Computer Engineering, The University of Texas, San Antonio, TX 78249. %Corresponding Author: Yongcan Cao (yongcan.cao@utsa.edu)
}
}

\begin{document}
\maketitle
\thispagestyle{empty}
\pagestyle{empty}

%%%%%%%%%%%%%%%%%%%%%%%%%%%%%%%%%%%%%%%%%%%%%%%%%%%%%%%%%%%%%%%%%%%%%%%%%%%%%%%%
\begin{abstract}
%\textcolor{red}{First two sentences are commented.}
Human demonstrations can provide trustful samples to train reinforcement learning algorithms for robots to learn complex behaviors in real-world environments. However, obtaining sufficient demonstrations may be impractical because many behaviors are difficult for humans to demonstrate. A more practical approach is to replace human demonstrations by human queries, \textit{i.e.,} preference-based reinforcement learning. One key limitation of the existing algorithms is the need for a significant amount of human queries because a large number of labeled data is needed to train neural networks for the approximation of a continuous, high-dimensional reward function. To reduce and minimize the need for human queries, we propose a new GAN-assisted human preference-based reinforcement learning approach that uses a generative adversarial network (GAN) to actively learn human preferences and then replace the role of human in assigning preferences. The adversarial neural network is simple and only has a binary output, hence requiring much less human queries to train. Moreover, a maximum entropy based reinforcement learning algorithm is designed to shape the loss towards the desired regions or away from the undesired regions. To show the effectiveness of the proposed approach, we present some studies on complex robotic tasks without access to the environment reward in a typical MuJoCo robot locomotion environment. The obtained results show our method can achieve a reduction of about $99.8\%$ human time without performance sacrifice.
\end{abstract}
\begin{keywords}
 Reinforcement Learning, Generative Adversarial Network (GAN), Human Preferences
\end{keywords}

%%%%%%%%%%%%%%%%%%%%%%%%%%%%%%%%%%%%%%%%%%%%%%%%%%%%%%%%%%%%%%%%%%%%%%%%%%%%%%%%
\section{Introduction} 

The application of reinforcement learning (RL) in solving complex problems has demonstrated success in domains with well-specified reward functions, (see, \textit{e.g.,} ~\cite{mnih2015human}). The existence of correct reward functions is crucial for the subsequent development of robotic control systems via reinforcement learning algorithms. However, designing such a reward function often requires the consideration of different objectives that can potentially impact not only the learned behaviors but also the learning process. This difficulty underlies some recent concerns about the misalignment between the reward functions and the objectives in the development of reinforcement learning algorithms~\cite{russell2016should}. Hence, it is crucial to communicate the actual objectives via properly defined rewards or reward functions, which often need to reflect humans' preferences since the mission objectives are often generated by humans.

To communicate humans' preferences effectively, it is important to develop methods and algorithms that can parse humans' intents and strategies. Two main approaches, namely, inverse reinforcement learning and human preference-based learning, have been proposed recently. Inverse reinforcement learning (see, \textit{e.g.,}~\cite{abbeel2004apprenticeship,ziebart2008maximum}) focuses on learning reward functions directly from human demonstrations. Human preference-based learning (see, \textit{e.g.,} ~\cite{eric2008active,dorsa2017active,palan2019learning}) focuses on maximizing the volume removed from the distribution of the weight vector by asking a human to pick between trajectory pairs until reaching convergence. While both approaches provide important advances, they still lack much-needed data efficiency. In particular, inverse reinforcement learning usually requires a large number of high-quality human demonstrations to obtain accurate reward functions given that the relationship between features of action-state pairs and the corresponding value in the form of reward is typically complex and difficult to be quantified. Nevertheless, obtaining a large volume of human demonstrations can be costly and even infeasible. Meanwhile, human preference-based learning also requires significant human feedback such that reward functions and/or control policies can be generated to produce robotic behaviors that meet humans' preferences~\cite{wirth2017survey}.
For example, in~\cite{christiano2017deep}, human queries are used to learn rewards, which are then used to train a reinforcement learning algorithm to generate trajectories. Although the proposed approach therein can potentially reduce human time in some existing human preference-based learning approaches, \textit{e.g.,}~\cite{furnkranz2012preference,akrour2012april}, a good amount of human queries are still needed to learn a high-dimensional (and often continuous) reward function. To practically design and train reinforcement learning algorithms with limited human feedback, it is essential to significantly decrease and minimize human feedback/efforts for efficiency and practicality.

\begin{figure}[t]
\centering
\includegraphics[width=1\columnwidth]{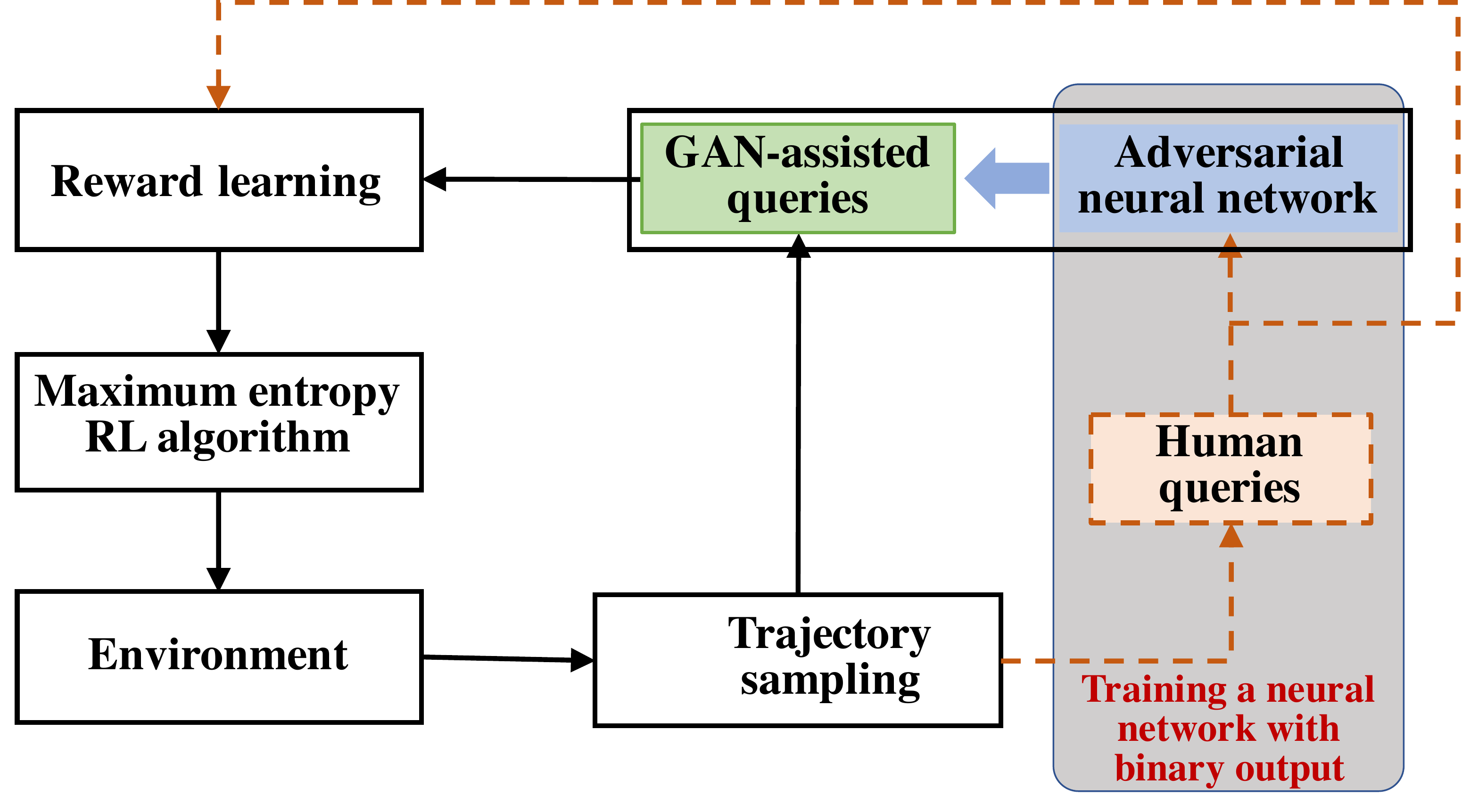} % Reduce the figure size so that it is slightly narrower than the column. Don't use precise values for figure width.This setup will avoid overfull boxes. 
\caption{The proposed GAN-assisted preference-based reinforcement learning framework. The dashed line means the flow in that branch is sparse.}
\label{fig1:frame}
\end{figure}

%three stage iterative framework to efficiently train PbRL policy that can dramatically decrease human feedback, e.g., reducing 77\% to 93\% human queries in the physics simulator MuJoCo~\cite{todorov2012mujoco}. 

In this paper, we propose a new GAN-assisted human preference-based reinforcement learning approach that can learn complex robotic behaviors with dramatically reduced human feedback. Fig.~\ref{fig1:frame} shows the overall structure of the proposed approach. In the proposed approach, pairs of queries are first sampled from the robot trajectories produced by the current policy and then sent to a human for selection of preferred ones. These (limited) human preference samples are used to train an adversarial neural network to learn human preference. The trained adversarial neural network is then used to replace human in the subsequent selection of preferred trajectories. The preference samples from humans and adversarial neural networks are used to learn reward functions. Then a maximum entropy based reinforcement learning algorithm is designed to effectively learn control policies that can generate robotic behaviors that meet human preferences. Instead of directly learning a high-dimensional reward function as in~\cite{christiano2017deep}, one key novelty of \textcolor{black}{our} approach is the design of a low-dimensional adversarial neural network to learn human preferences. Due to the low-dimensional nature of the adversarial neural network, much less human queries are needed in the training process. Another key advantage of the proposed GAN-assisted human preference reinforcement learning approach is that one set of human queries \textcolor{black}{can serve two different purposes, namely, (\textit{i}) training a GAN that directly learns the principles of human preference, and (\textit{ii}) learning reward functions for the control policy generation.} Moreover, the construction of adversarial neural network does not depend on the reward function learning process. By sending queries to the trained adversarial neural network, rather than humans, in the reward learning and control policy generation process, more significant human time reduction can be achieved without performance sacrifice. %Without creating such an adversarial neural network, the existing approach, focusing on directly training a good reward function based on human queries to produce actions that label trajectory pairs, will require more human queries due to the need for learning a high-dimensional reward function.

This paper has three main contributions. First, we create a new GAN-assisted human preference-based reinforcement learning approach that trains a low-dimensional adversarial neural network to learn and predict human preferences based on very limited human queries. The trained adversarial neural network can \textcolor{black}{then replace the role of humans} in \textcolor{black}{labeling the trajectory pairs, which can dramatically reduce human labeling time}. Second, we propose a maximum entropy reinforcement learning algorithm using a new nonlinear objective function based on human preferences when human preferences among trajectory pairs are used to shape the loss towards the desired regions or away from the undesired regions. Third, we conduct experiments on complex robotic tasks to show that the proposed approach can outperform the state-of-the-art methods. In particular, the proposed approach can further reduce 84\% human time needed in~\cite{christiano2017deep} while achieving better performance. Overall, the proposed method can reduce approximately $99.8\%$ human time without performance sacrifice. 

%The remainder of the paper is organized as follows. Section~\ref{sec:re_wo} provides an overview of some relevant research as the background knowledge. Then we will present the main technical approach in Section~\ref{sec:TA}. Section~\ref{sec:exp} shows the experimental results and their comparison with the existing state-of-the-art results. Finally, Section~\ref{sec:con} briefly summarizes the main contribution of our work.

%the human time and the  when synthetic feedback with $4.5\times$ amount of labels is used. This implies a reduction of 77\% to 93\% queries without performance loss. 

%Direct learn a policy to generate human preferences

%Maximum entropy based preference based learning with reduced queries

%include adversarial network to provide preferences to further shape the reward, even correct humans' preferences...

\section{Related Work} \label{sec:re_wo}
Some interesting studies have been conducted on the development of reinforcement learning algorithms based on human ratings or rankings (see, \textit{e.g.,}~\cite{akrour2012april,akrour2014programming,daniel2015active,el2016score,wang2016learning} and~\cite{wirth2016model}). Additional interesting research can be found in, \textit{e.g.,}~\cite{furnkranz2012preference}, which focuses on the general problem of reinforcement learning from preferences instead of absolute reward values. In~\cite{Yang2019NoRMLNM}, the research focus is on tuning the task specific policy using observable dynamics of the environment instead of an explicit reward function. In~\cite{sorensen2016breeding,palan2019learning}, the research focus is on optimization using human preferences in settings other than reinforcement learning.

Integrating human demonstrations and preferences fits a recent trend of communicating complex personalized objectives to deep learning systems in, \textit{e.g.,} inverse reinforcement learning~\cite{dorsa2017active,biyik2018batch,fu2018variational}, imitation learning~\cite{stadie2017third}, guided cost learning~\cite{finn2016guided}, and semi-supervised skill generalization~\cite{finn2016generalizing}. For example, \cite{dorsa2017active} assumed that the reward function is parameterized by a linear function of hand-coded features. First, a probabilistic model was implemented to capture the preference. Then the Bayesian inference approach was developed to fit the reward function. Finally, a constrained optimization problem was formulated to solve the trajectory selection problem without using reinforcement learning based methods. One key challenge in the approaches proposed in the aforementioned references is the consideration of complex robotic tasks while addressing human preferences specified via limited actual human queries.

In~\cite{singh2019end,macglashan2017interactive}, experiments were performed that involve reinforcement learning from actual human feedback. In~\cite{singh2019end}, similar to the standard reward-based reinforcement learning, a classifier was first trained on an initial set of trajectories. Then a binary variable is used to cast success as the reward. In the training process, only the state with the largest value for success will be selected for query. This labeled state is added to the trajectories for the fine-tuning of the classifier. Note that the proposed method in~\cite{singh2019end}, similar to that in~\cite{christiano2017deep}, does not include the active learning of human preferences. In other words, no model is trained to predict and replace human queries. Hence, the required human time is still high (1-4 hours).  

Our work focuses on improving the sample efficiency of reinforcement learning algorithms with human preferences when learning complex robotic behaviors. Although the proposed approach has a similar structure as~\cite{christiano2017deep}, there are \textcolor{black}{two} fundamental differences. First, \textcolor{black}{we design an adversarial neural network that directly learns human preference while \cite{christiano2017deep} does not include the direct human preference learning process}. The trained adversarial neural network can replace humans in the labeling of trajectory pairs, hence dramatically reducing human time. Because the adversarial neural network only outputs preferences rather than rewards, very limited human queries \textcolor{black}{are} required. Second, a standard reinforcement learning algorithm is used in~\cite{christiano2017deep}, which requires longer training time with a higher chance of finding locally optimal control policies \textcolor{black}{(a common issue in reinforcement learning~\cite{haarnoja2017reinforcement})}. We adopt a maximum entropy based reinforcement learning algorithm with a newly designed nonlinear objective function, which can shape the loss towards the desired regions or away from the undesired regions. Hence, the proposed approach can reduce \textcolor{black}{human/GAN-assisted} queries while improving performances.  %approach of \cite{akrour2012april,akrour2014programming} and \cite{christiano2017deep}  \cite{christiano2017deep} focuses on scaling human feedback up to deep reinforcement learning and learning much more complex behaviors. They considered physics tasks with dozens of degrees of freedom and Atari tasks with no hand-engineered features. 
% GAN

%Since the adversarial neural network is to provide a binary output rather than a high-dimensional reward function, it is expected to be less sensitive to inconsistent human queries than the reward learning in~\cite{christiano2017deep}. Because humans could make mistakes when labeling the trajectory pairs, especially when taking long-term tasks, it is desirable to develop a more robust model. Moreover, the adversarial neural network, once trained, can be used to correct incorrect/inconsistent labels by humans, which cannot be achieved using the existing methods in~\cite{christiano2017deep,singh2019end}. Since our focus here is not on the understanding of the robustness of the proposed approach, we will report further results later.

\section{Technical Approach} \label{sec:TA}

%the process of an agent, namely human or GAN, interacting with

%\subsection{The Proposed Framework}
In this section, we will provide details about the proposed GAN-assisted preference-based reinforcement learning approach. For the purpose of generality, we consider a general environment modeled as a Markov decision process (MDP), defined by the tuple $({S},{A}, {T},r,\gamma,q_0)$, where ${S}$ is the (continuous) state space, ${A}$ is the (continuous) action space, ${T}$ is the dynamics or transition distribution, $\gamma \in (0, 1]$ is the discount factor, $r$ is the reward function, and $q_0$ is the initial state distribution.
We further use $\rho_\pi(s,a)$ to denote the state-action marginal of the trajectory distribution induced by a policy $\pi(a|s)$.%, and  \textcolor{yellow}{$\rho(s,s',a):  {S} \times  {S} \times  {A} \rightarrow [0,\infty)$ to represent the state transition probability density, namely, the probability density of the next state $s'$ given the current state $s$ and action $a$.} \textcolor{red}{I think this notation is already included in $T$.}

Instead of assuming that the environment directly produces a reward, we consider the case when there exists a human who observes and expresses his/her preferences between pairs of trajectory segments $(\tau_1,\tau_2)$ produced based on the current policy $\pi$. The trajectory segment is typically of the form \textcolor{black}{$(s_i,a_i,s_{i+1},a_{i+1},\cdots, s_{i+k})$}. At each time step, the proposed method includes the construction of three components: (1) a control policy $\pi:  {S} \rightarrow  {A}$; (2) an adversarial network that prefers one trajectory (say, $\tau_1$) to another (say, $\tau_2$), whose quantitative representation is given by $p(\tau_1 \succ \tau_2)$, where $p(\tau_1 \succ \tau_2)$ denotes the probability of selecting $\tau_1$ not $\tau_2$ by the human; and (3) an estimated reward function $\hat{r}(s,a):  {S} \times  {A} \rightarrow  {R}$. 

Note that all three components involve the construction and update of deep neural networks based on the interaction with the environment. In particular, we propose to update the three networks iteratively via the following steps:
\begin{itemize}
\item[1.] Send pairs of segments $(\tau_1,\tau_2)$ produced via the current control policy $\pi$ as human queries;
\item[2.] Update an adversarial neural network, projecting trajectories $\tau_i$ to their preferences $\xi$, to learn human preference, where $\xi$ is a distribution over $\left\{1, 2\right\}$ indicating which segment is preferred; 
\item[3.] Train a reward model to predict the human preference-based Q value;
\item[4.] Update parameters to obtain a new control policy by a maximum entropy based reinforcement learning algorithm to maximize the human preference-based Q value.
\end{itemize}
The four steps run iteratively until the three neural networks produce stable outcomes, namely, the difference between two consecutive steps being smaller than a threshold. More technical details are provided in the following subsections.
%with human comparisons flowing from stage (1) to stage (2), policy optimization flowing stage (2) to stage (3), and trajectories flowing from stage (3) to stage (1). The following subsections provide details on each of these processes.

\subsection{Preference Queries}
At each time step $t$, a robot receives an observation $s_t \in  {S}$ from the environment and then generates an action $a_t \in  {A}$ for the robot to interact with the environment. 
In the policy evaluation step, the Q value can be computed iteratively, starting from any function $Q: {S}\times {A}\rightarrow {R}$ and repeatedly applying a modified Bellman backup operator $\psi^\pi$ given by~\cite{haarnoja2017reinforcement} 
$$[\psi^\pi Q](s_t,a_t)\triangleq \hat{r}(s_t,a_t) + \gamma  {E}_{s_{t+1}\sim\rho_\pi}[V(s_{t+1})],$$
where $V(s_t)= {E}_{a_t\sim\pi}[Q(s_t,a_t)-\log\pi(a_t|s_t)]$ and $ {E}$ is the expectation operator. The agent's goal is to maximize the discounted sum of the \textcolor{black}{reward} values. \textcolor{black}{Because the environment may not produce a reward signal,} we assume that the human can provide preferences between trajectory segments. \textcolor{black}{In particular,} for a pair of trajectory segments $\tau_1$ and $\tau_2$, we follow the definition of preference over trajectories in~\cite{wirth2017survey}. More specifically, the notation $\tau_1 \succ \tau_2$ means that the first choice is {strictly preferred}. The notation $\tau_2 \succ \tau_1$ means that the second choice is {strictly preferred}. The notation $\tau_1 \sim \tau_2$ means that the two choices are {indifferent}, \textit{i.e.}, neither $\tau_1 \succ \tau_2$ nor $\tau_2 \succ \tau_1$ holds. 
The goal of the robot is to produce trajectories that are preferred by the human, while \textcolor{black}{requiring} as few human queries as possible. 

We here define the criteria for preferences over trajectories in three ways:
\begin{itemize}
\item[C1.] Start from a goal expressed in natural language, ask the {human} to evaluate the behavior, and then give a binary feedback to indicate whether $\tau_1 \succ \tau_2$.
\item[C2.] Start from a goal expressed in natural language, ask the {adversarial neural network} to evaluate the behavior, and then give a binary feedback to indicate if $\tau_1 \succ \tau_2$.
\item[C3.] For a given estimated reward function $\hat{r}:  {S} \times  {A} \rightarrow  {R}$, the preference yields $\tau_1 \succ \tau_2$ if $\sum_{i=1}^k \textcolor{black}{\gamma^{i-1}}\hat{r}(s_{t+i}^1,a_{t+i}^1) >\sum_{i=1}^k \textcolor{black}{\gamma^{i-1}}\hat{r}(s_{t+i}^2,a_{t+i}^2),$ where $\tau_1=\{(s_{t+i}^1,a_{t+i}^1),~i=1,\cdots,k\}$ and $\tau_2=\{(s_{t+i}^2,a_{t+i}^2),~i=1,\cdots,k\}$, and $\tau_2 \succ \tau_1$ otherwise. 
\end{itemize}
%In other words, a preference $\tau_1 \succ \tau_2$ is satisfied if 
%\begin{equation}
%\tau_1 \succ \tau_2 \Leftrightarrow Pr_\pi(\tau_1) > Pr_\pi(\tau_2),\nonumber
%\end{equation}
%where
%$
%Pr_\pi(\tau) = q_0(s_0)\sum_{t=0}^{\left|\tau\right|}\pi(a_t|s_t) {T}(s_{t+1}^{\prime}|a_t,s_t)\nonumber
%$
%is the probability of obtaining a trajectory $\tau$ with the policy $\pi$.
C1 and C3 were proposed in~\cite{christiano2017deep} to generate references quantitatively. In particular, C1 uses direct human queries to obtain references, while C3 seeks to use the learned reward to generate references based on their accumulated rewards. Because C3 involves two steps of approximation, namely, approximation of both the reward and the policy, it requires a good reward model in order to yield high-quality trajectory selection subject to human preferences. In addition, the reward function to be learned is typically high-dimensional. In contrast, the proposed C2 seeks to learn the binary human preferences directly, namely, $\tau_1 \succ \tau_2$ or $\tau_2 \succ \tau_1$, hence requiring less data to train a good model that can predict binary human references. In particular, the proposed C2 method is motivated by the construction of generative models to directly generate samples in GAN~\cite{goodfellow2014generative}. Instead of generating close-to-true (yet fake) samples in GAN, the proposed C2 seeks to generate a close-to-true prediction of binary human preferences. %In addition, C2 and C3 may be used concurrently, without human queries, to provide preference learning from multiple views for improved accuracy with less human queries.

In the process of preference queries, two trajectory segments in the form of short video clips are provided to the human or the adversarial neural network for labeling.\footnote{In our experiments, these clips have a duration of $5$ seconds.} In particular, these video clips will be sent to the human at the beginning and the adversarial neural network later once the trained adversarial neural network model, via using the human selection samples, can provide a very high accuracy of human selections. Afterwards, the human or the adversarial neural network indicates which segment is preferred, or the two segments are indifferent. The outcomes are stored in a database $D$ \textcolor{black}{in the form of a} tuple $(\tau_1, \tau_2, \xi)$, where $\tau_1$ and $\tau_2$ are the two clips and $\xi$ is a \textcolor{black}{discrete} distribution over $\left\{1, 2\right\}$ indicating which clip the human or the adversarial neural network prefers. In particular, if one clip is preferred, $\xi$ puts all of its mass on that choice. If the clips are equally preferable, then $\xi$ is uniform. Moreover, the human has one more option, \textit{i.e.}, marking the trajectory segments as incomparable. In this case, the comparison will not be included in the database $D$.

\subsection{Training Adversarial Neural Network}
We now move to the discussion of training the adversarial neural network in \textcolor{black}{criterion} C2. The objective of an adversarial neural network is to create a model that can {actively} learn human preference so that human queries can be replaced by adversarial neural network queries. Moreover, the trained adversarial neural network can be used to correct human labels if the human provides incorrect labels due to, \textit{e.g.,} fatigue. We here propose to train a discriminative neural network model $Y_{di}$, parameterized by $\theta_{di}$, that can learn the probability of selecting a sample trajectory by the human. Let $D_{p}$ denote the preferred trajectory set selected from the sampled trajectories set $D$, and $D_{np}$ denote $D \setminus D_{p}$, \textit{i.e.,} the non-preferred trajectory set. The model parameter $\theta_{di}$ can be updated by the stochastic gradient ascent algorithm as
\begin{align}\label{equ:ann}
\theta_{di}\leftarrow~ &\theta_{di}+\nabla_{\theta_{di}}  {E}_{\tau_i \in D_{np}}\log(1- Y_{di}(\tau_i)) \notag\\
& + \nabla_{\theta_{di}}  {E}_{\tau_i \in D_{p}}\log Y_{di}(\tau_i).
\end{align}

%\textcolor{red}{Based on my understanding, the GAN only plays a role of classifier (Discriminator), we may need to think why not using a binary classifier instead?}

\subsection{Training Reward Function}
Since the environment \textcolor{black}{reward is unavailable, we here propose to leverage the GAN-assisted human preference to approximate the reward function.} Human preference is needed at the very beginning to produce reasonable data points to train the adversarial neural network \textcolor{black}{and} the adversarial neural network can be used to replace human queries \textcolor{black}{once} it provides a good approximation of human preferences. 

In the case of human preference, we can interpret a preference-based Q value function estimation $Q(s_t,a_t)$ as a preference predictor if we view $\hat{r}$ as a latent factor explaining the human's judgments. In other words, we have $Q(s_t,a_t)=\sum_i \hat{r}(s_{t+i},a_{t+i})$. By following~\cite{dorsa2017active}, we here assume that the human's probability of preferring a clip $\tau_i$ depends exponentially on the value of the latent reward summed over the length of the clip. Then we can \textcolor{black}{calculate} the probability $\xi$ as capturing the preference with respect to $Q(s_t,a_t)$, \textcolor{black}{such that,}
\begin{equation} \label{eq: cal_preference}
\begin{split}
\hat{p}(\tau_1 \succ \tau_2) &= \frac{1}{1+\exp(Q(s^2_t,a^2_t)-Q(s^1_t,a^1_t))}, \\
 \hat{p}(\tau_2 \succ \tau_1) &= \frac{1}{1+\exp(Q(s^1_t,a^1_t)-Q(s^2_t,a^2_t))}, 
\end{split}
\end{equation}
where $\hat{p}(\tau_1 \succ \tau_2)$ means the estimated probability of selecting $\tau_1$ not $\tau_2$, namely, ${p}(\tau_1 \succ \tau_2)$, and $\hat{p}(\tau_2 \succ \tau_1)$ means the estimated probability of selecting $\tau_2$ not $\tau_1$, namely, ${p}(\tau_2 \succ \tau_1)$. Following the idea in~\cite{christiano2017deep}, we can \textcolor{black}{update the Q value function} to minimize the cross-entropy loss ${L}_{a}$ between the \textcolor{black}{the calculated human preference-based labels from (\ref{eq: cal_preference})} and the actual human labels as
\begin{align}\label{loss:ho}
 {L}_{a} = - \sum_{(\tau_1,\tau_2,\xi)\in D} [\xi(1)& \log \hat{p}(\tau_1 \succ \tau_2) + \xi(2) \log \hat{p}(\tau_2 \succ \tau_1)].
\end{align}

In the adversarial neural network testing, the probability $\xi$ is modeled by
\begin{align}
p(\tau_1 \succ \tau_2) = Y_{di}(\tau_1),\quad
p(\tau_2 \succ \tau_1) = Y_{di}(\tau_2). 
\end{align}
Note that $p(\tau_1 \succ \tau_2) =  1 - p(\tau_2 \succ \tau_1)$ does not hold in general, \textcolor{black}{as the two values are generated from the adversarial neural network independently. Therefore,} we here propose to minimize the cross-entropy loss $ {L}_{g}$ between the predicted human preference-based labels and the actual GAN-assisted labels given by
\begin{align}\label{loss:g}
 {L}_{g} =& - \sum_{(\tau_1,\tau_2,\xi)\in D} \Big[p(\tau_1 \succ \tau_2) \log \hat{p}(\tau_1 \succ \tau_2) \nonumber\\ 
&~~~~~~~~~~~~~~~~+ (1 - p(\tau_1 \succ \tau_2)) \log \hat{p}(\tau_2 \succ \tau_1)\Big]\nonumber \\
&- \sum_{(\tau_1,\tau_2,\xi)\in D}\Big[ p(\tau_2 \succ \tau_1) \log \hat{p}(\tau_2 \succ \tau_1) \nonumber\\ 
&~~~~~~~~~~~~~~~~+ (1 - p(\tau_2 \succ \tau_1)) \log \hat{p}(\tau_1 \succ \tau_2)\Big].
\end{align}
The difference in predicted reward of two trajectory clips estimates the probability that one is chosen over the other due to the (non-expert) human preference in~\eqref{loss:ho} or the GAN-assisted human preference in~\eqref{loss:g}. A higher reward for a trajectory clip means a higher probability to be selected.

\subsection{Training a Maximum Entropy Reinforcement Learning Algorithm}

After obtaining the estimated reward $\hat{r}$, we are left with the design of a reinforcement learning algorithm for control policy generation. Because the reward function $\hat{r}$ may be non-stationary and a sample-efficient deep reinforcement learning algorithm is preferred, we here focus on a maximum entropy reinforcement learning problem that combines off-policy updates with a stable stochastic actor-critic formulation. 

%In this section, we will formally show the equivalence between the maximum entropy RL algorithm and a GAN in which the generator's density can be evaluated and provided as an additional input to the discriminator. Hence, the only hyperparameter that needs to be adjusted is the temperature parameter $\alpha$, shown in~\eqref{eq:alpha}, which determines the relative importance of the entropy term against the reward. We normalize the rewards such that they have zero mean and a constant standard deviation, also called reward normalization. This is a typical preprocessing step since the value of the rewards is to be determined.

%Assume that $\int _ {A}\exp(\frac{1}{a}Q_{soft}(\cdot,a^{\prime}))da^{\prime}< \infty$ exists. 

%\begin{theorem}\label{theory:loss}
%Consider a map $Q: {S}\times  {A}\rightarrow  {R}$ with $\left |  {A} \right |<\infty$, and define $Q^{k+1} = \psi^\pi Q^k$. Then $ {L(\theta)} =  {L}(D_{NPr})$, where $ {L(\theta)} $ is the loss of the PbRL objective and $ {L}(D_{NPr})$ is the loss of the generator.
%\end{theorem}
%\begin{proof} 

Since the objective is to learn the optimal policy $\pi^\ast$, the maximum entropy reinforcement learning problem can be written as
\begin{equation}\label{eq:maxent}
\pi^{\ast} = \argmax_\pi \sum_{t} \textcolor{black}{\gamma^{t}}  {E}_{(s_t,a_t) \sim \rho_\pi} r^{ME}(s_t,a_t,\pi),
\end{equation}
where $r^{ME}(s_t,a_t,\pi)\triangleq \hat{r}(s_t,a_t) + \alpha  {H}(\pi(\cdot|s_t))$ is a more general maximum entropy reward in the maximum entropy reinforcement learning setting~\cite{haarnoja2017reinforcement} that favors stochastic policies, and $\alpha$ is a parameter that determines the relative importance of entropy $ {H}(\pi(\cdot|s_t))$ \textcolor{black}{with respect to the} reward ${r}(s_t,a_t)$.
Based on the structure in~\eqref{eq:maxent}, we now define a new preference-based policy learning problem as
\begin{align}\label{eq:pbRLPolicy}
{\pi}^{\ast} = \argmax_\pi &\sum_l \Bigg\{ {E}_{D_{p}} [\gamma^lr^{ME}(s_{t+l},a_{t+l},\pi)] \notag\\
&-  {E}_{D_{np}}\left. \log \left(\frac{\exp [\gamma^lr^{ME}(s_{t+l},a_{t+l},\pi))]}{\pi(a_{t+l}|s_{t+l})}\right)\right\}.
 \end{align} 
Compared to~\eqref{eq:maxent}, the new formulation in~\eqref{eq:pbRLPolicy} has a new $\log$ term that is intentionally added to reflect humans' preference/desire. The new formulation allows us to learn a nonlinear objective function based on human preferences because human preferences among trajectory pairs can shape the loss towards the desired regions and away from the undesired regions. Hence, optimal policies can be learned from the objective function. %We will provide more intuition behind this policy in~\eqref{eq:gradient}. 

Under the actor-critic setting, let the policy $\pi$ be parameterized by $\theta$ via an actor network. Then the objective function in~\eqref{eq:pbRLPolicy} can be written as
\begin{align}\label{eq:MaxEntpr}
 {L}(\theta) =&\sum_l \Bigg\{ {E}_{D_{p}} [\gamma^lr^{ME}(s_{t+l},a_{t+l},\pi_\theta)] \notag\\
&-  {E}_{D_{np}} \log \left. \left(\frac{\exp [\gamma^lr^{ME}(s_{t+l},a_{t+l},\pi_\theta))]}{\pi_\theta(a_{t+l}|s_{t+l})}\right)\right\},
\end{align}
where $\pi_\theta$ denotes the parameterized $\pi$.
%where $D$ denotes the trajectory set sampled from the current policy, $D_{Pr}$ denotes the prefered trajectories set selected from the sampled trajectories set $D$, and $D_{NPr}$ denotes $D-D_{Pr}$.
Let $q = \sum_i w_i$ and
$$w_i = \frac{\exp [\gamma^lr^{ME}(s_{t+l},a_{t+l},\pi_\theta))]}{\pi_\theta(a_{t+l}|s_{t+l})}.$$ The gradient of $ {L}(\theta)$ is given by
\begin{align}\label{eq:gradient}
\frac{d {L}}{d\theta} = & {E}_{D_{p}}\sum_{l}\frac{d(\gamma^lr^{ME}(s_{t+l},a_{t+l},\pi_\theta))}{d\theta}\nonumber\\& -  {E}_{D_{np}}\frac{1}{q} \sum_{l}w_i\frac{d(\gamma^lr^{ME}(s_{t+l},a_{t+l},\pi_\theta))}{d\theta}.
\end{align}  

From the gradient computation in~\eqref{eq:gradient}, we can get some insights of the policy derived from~\eqref{eq:pbRLPolicy}. First, the magnitude of the maximum entropy reward directly contributes to the gradient ascent while the human preferred trajectories can shape the loss towards the desired regions and the human unpreferred trajectories can shape the loss away from undesired regions. Second, the denominator $\pi(a_{t+l}|s_{t+l})$ in~\eqref{eq:pbRLPolicy} denotes the background distribution from which the trajectories $\tau_i$ are sampled using the current policy $\pi_\theta$. %Hence the distribution is normalized.

%We next show that the loss function~\eqref{eq:MaxEntpr} can be equivalently written as a generator in GAN. For a fixed generator with an density $\hat{\pi}$, the optimal discriminator is given by~\cite{goodfellow2014generative}
%\begin{align}
%D_\theta(\tau)=\frac{p_\theta(\tau)}{p_\theta(\tau)+\hat{\pi}(\tau)}.
%\end{align}
%Where $p_\theta(\tau)$ refers to the probability with respect to a preferred trajectory. According to~\cite{christiano2017deep}, 
%the human selects preferred trajectories with a probability $p^{Pr}_\theta(\tau)$ that can be modeled as 
%\begin{align}
%p_\theta(\tau) =  \frac{\sum_l \gamma^lr^{ME}(s_{t+l},a_{t+l},\pi_\theta)}{\sum_{\tau\sim\rho_\pi}\sum_l \gamma^lr^{ME}(s_{t+l},a_{t+l},\pi_\theta)}.
%\end{align}
%According to~\cite{goodfellow2014generative}, the generator's loss can be computed by
%\begin{align}\label{eq:GAN}
%& {L}(D_{samp})= {E}_{\tau\sim D}[\log(1-D_\theta(\tau))-\log(D_\theta(\tau))]\nonumber \\
%&=E_{\tau \sim D_{samp}}\left[\log\frac{\pi(\tau)}{p^{Pr}_\theta(\tau)+\pi(\tau)}-\log\frac{ \frac{ \exp(Q^{\theta}_{soft}(s_t,a_t))}{Z}}{p^{Pr}_\theta(\tau)+\pi(\tau)}\right]\nonumber\\
%&=E_{\tau \sim D_{samp}}\left[\log\pi(\tau)-  Q^{\theta}_{soft}(s_t,a_t)+log(Z)\right],
%\end{align}
%where $ Z = \int_{s_0, a_0,\cdots\sim\pi_\theta}\sum_{t} \exp(\frac{1}{\alpha}r(s_t,a_t))$. 

%Note that the loss in~\eqref{eq:GAN} is the same as the preference-based objective function, defined in~\ref{eq:MaxEntpr}. This completes the proof of the theorem.

\subsection{Pseudocode of the Overall Approach}

The pseudocode for the proposed GAN-assisted human preference-based reinforcement learning approach is shown in Algorithm~\ref{alg:PbRL}.

\begin{algorithm}[ht]
\caption{Pseudocode for the proposed GAN-assisted human preference-based reinforcement learning approach}
\label{alg:PbRL}
\KwData{initial state distribution $q_0(s)$; initial policy  $\pi_{0}$; iteration limit  $m$; state sample limit  $n$; rollout limit  $k$; time step size for fine-tuning adversarial neural network  $T$.}
\KwResult{improved policy: $\pi_i$}
\emph{$ {A}(s)$: \textup{the set of actions available in states}}\\
\For{$ i  = 1$ \rm to $m$}{
\For{$0$ \rm to $n$}{
$s \sim$ $q_0(s)$

$\uptau = \varnothing ,\xi = \varnothing$

\For{$\forall a \in  {A}(s) $}{
\For{$0$ \rm to $k$}{
$\uptau \leftarrow \uptau~\cup$ \rm \textsc{rollout}$(s,a,\pi_i)$}}
\If{\rm \textsc {obtain~preference~via~C2} \textbf{and} \sc pathback($T$)}{
$~D_p, D_{np} \leftarrow$ C1 \rm or C3\\
\rm \textsc{fine-tune}($\uptau, D_p, D_{np}$) by maximizing~\eqref{equ:ann} via stochastic gradient ascent }
$\xi \leftarrow$ \rm \textsc{obtain~preference~via~C2}}
$~\pi_i \leftarrow $ \textsc{update~policy} by maximizing~\eqref{eq:MaxEntpr}\\
$Q_0(s,a)\leftarrow Q_0(s,\arg\max_aQ_0(s,a))$ by minimizing~\eqref{loss:g} via stochastic gradient descent
}
\end{algorithm}

\section{Experiments} \label{sec:exp}
%\subsection{Setup}
In this section, we will present some experimental results to evaluate the performance and effectiveness of the proposed method. All experiments are conducted on the MuJoCo physics engine~\cite{todorov2012mujoco}. For the \textcolor{black}{actor network}, we adopt a simple convolutional neural network with two hidden layers of size $64$ using ReLU~\cite{ramachandran2017searching} as the activation function. The discount factor is selected as $\gamma = 0.99$. The critic network has one hidden layer of size $64$. Both the adversarial neural network and the neural network that estimates the human-preference based Q values have two hidden layers of size $64$. We use the asynchronous advantage actor-critic (A3C) setting~\cite{babaeizadeh2016reinforcement} and execute parallel episodes in one batch. For all experiments, the number of environment copies is selected as $6$. The parameters are optimized using the stochastic gradient descent (SGD) algorithm~\cite{zhang2015deep} and a learning rate of $3 \times 10^{-4}$. All experiments are performed using TensorFlow, which supports automatic differentiation through the gradient updates~\cite{abadi2016tensorflow}.

%\subsection{Simulated Robotics}
In the following part of the section, we will demonstrate how robotic behavior learning tasks are solved using the proposed GAN-assisted preference-based reinforcement learning approach without accessing the true reward. In the setting of human queries, the human is first given a natural language description of each task and then provided some examples of labeling the trajectory clips based on the description. Then the human is asked to compare non-labeled clip pairs as the feedback. In particular, we consider two types of human queries, namely, human-$175$ and human-$345$. In particular, human-$175$ (respectively, human-$345$) asks the human to first label $175$ clips (respectively, $345$ clips) and then only label $6$ online clips per episode during the training process for each task. Each trajectory segment has a duration of $5$ seconds. Typically, the human can respond to the query in $5$ seconds. Since the clips are automatically generated without the need for human supervision, the human time is \textcolor{black}{only} calculated based on the approximated human labeling time under the assumption that each query takes $5$ seconds. Under this setting, typical experiments involving direct human queries requires less than $1$ hour of human time.

For comparison, we also run experiments using a synthetic (quantitative) oracle whose preferences over trajectory pairs exactly reflect accumulated reward in each specific task. In our quantitative oracle, instead of sending the queries to the human, the feedback is labeled by selecting clips receiving a higher reward in that task via the trained maximum entropy reinforcement learning algorithm in Section~\ref{sec:TA}.E. In the setting of synthetic queries, we consider the case synthetic-175 when $175$ clips were first labeled by the trained maximum entropy reinforcement learning algorithm and $175$ more online clips were labeled for every $100$ episodes during the training of the reinforcement learning algorithm. 

Finally, we also run experiments using human preferences learned by the adversarial neural network in Section~\ref{sec:TA}.C. In particular, the GAN-assisted human preferences are obtained via a pre-trained adversarial neural network based on learning the human's preferences over clip pairs. Instead of labeling the queries using the trained maximum entropy reinforcement learning algorithm, the adversarial neural network can directly label the clip pairs since it aims to produce a close-to-true prediction of human preferences. In the setting of GAN-assisted queries, we consider two types of GAN-assisted queries, namely, GAN-assisted-50 and GAN-assisted-175. Specifically, GAN-assisted-50 (respectively, GAN-assisted-175) asks the adversarial neural network to first label $50$ (respectively, $175$) video clips and then labels $50$ (respectively, $175$) online clips for every $100$ episodes during the training of the reinforcement learning algorithm. 

%The proposed method can save from ${5.09}$ hours to ${17.80}$ hours of human time.

%We also report the baseline of RL training using the real reward. The result is shown in Figure~\ref{fig:reward}.

\subsection{Results and Comparison}

\begin{figure*}[ht]
\centering
\subfloat{\includegraphics[height=5cm]{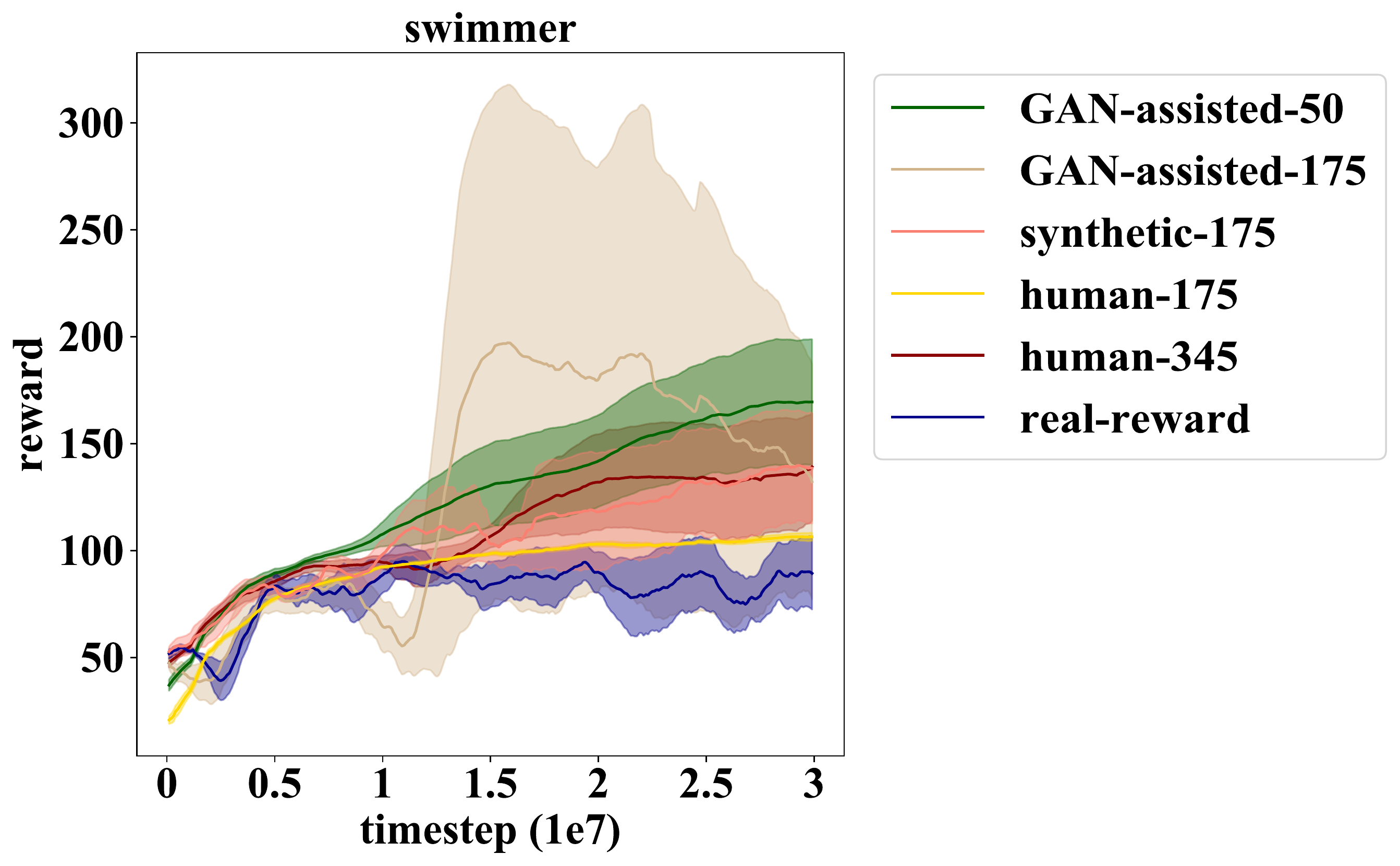}}~~
\subfloat{\includegraphics[height=5cm]{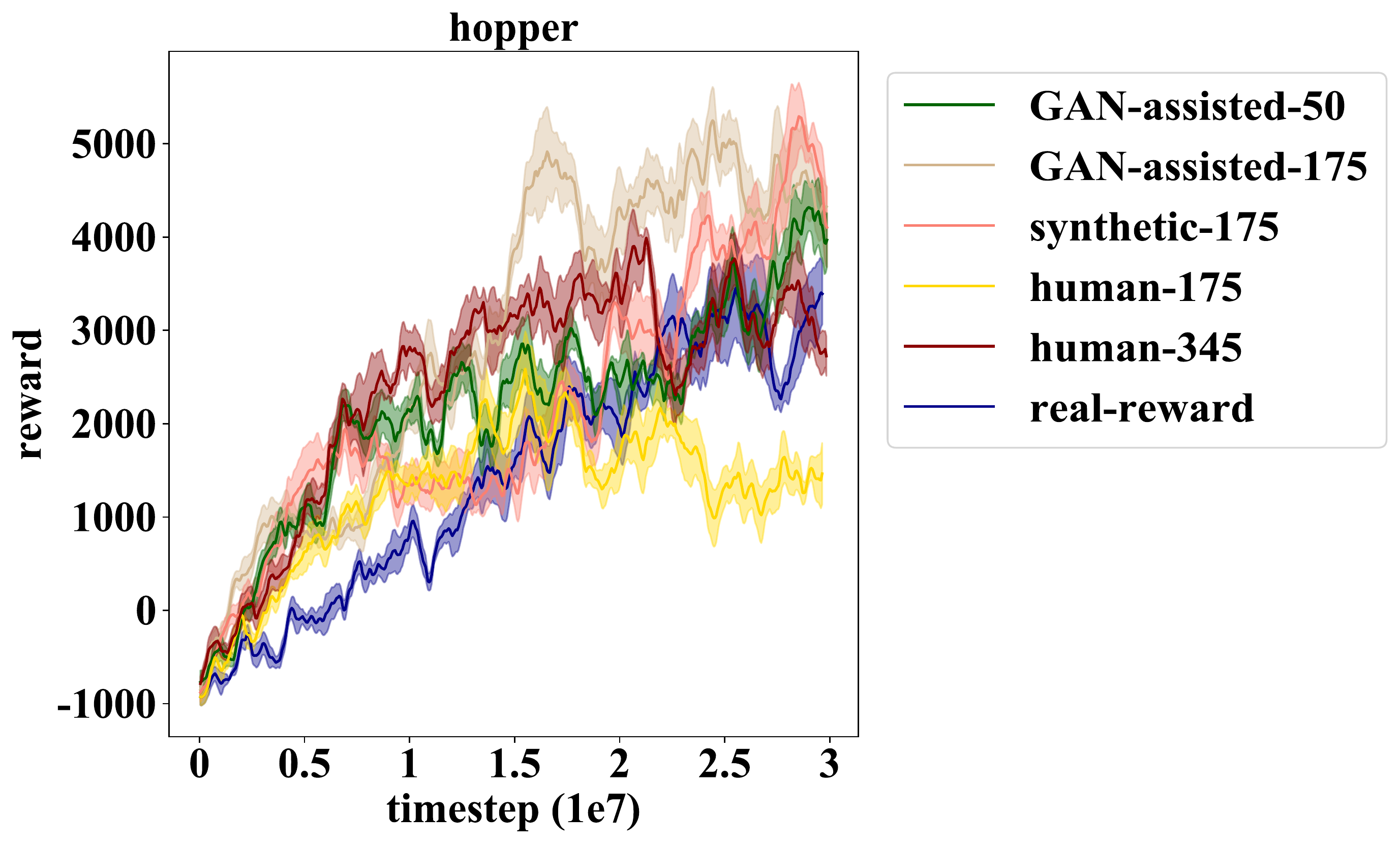}}\\
\subfloat{\includegraphics[height=5cm]{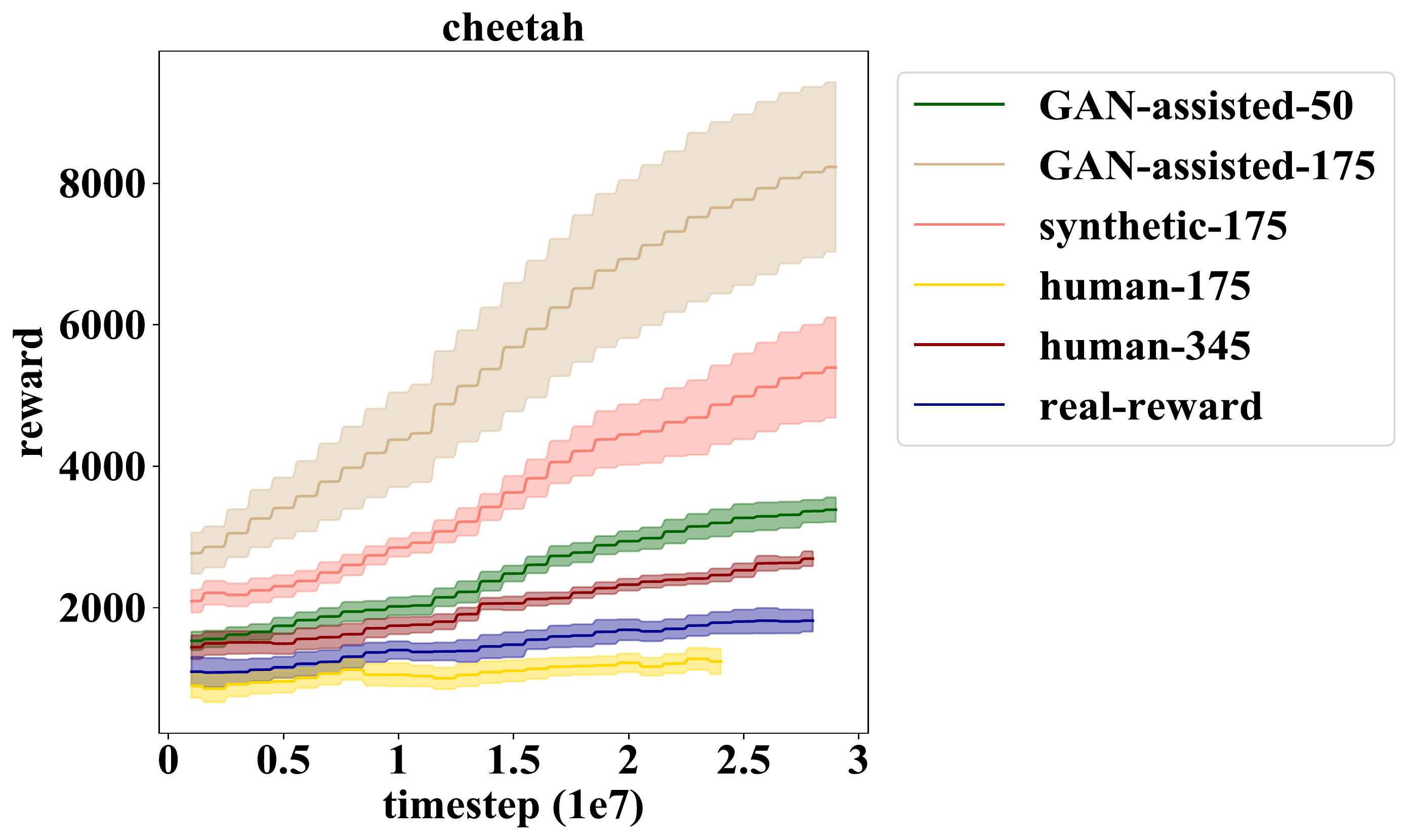}}~~
%\subfloat{\includegraphics[height=6cm]{./image/pendulum.pdf}}\\
\subfloat{\includegraphics[height=5cm]{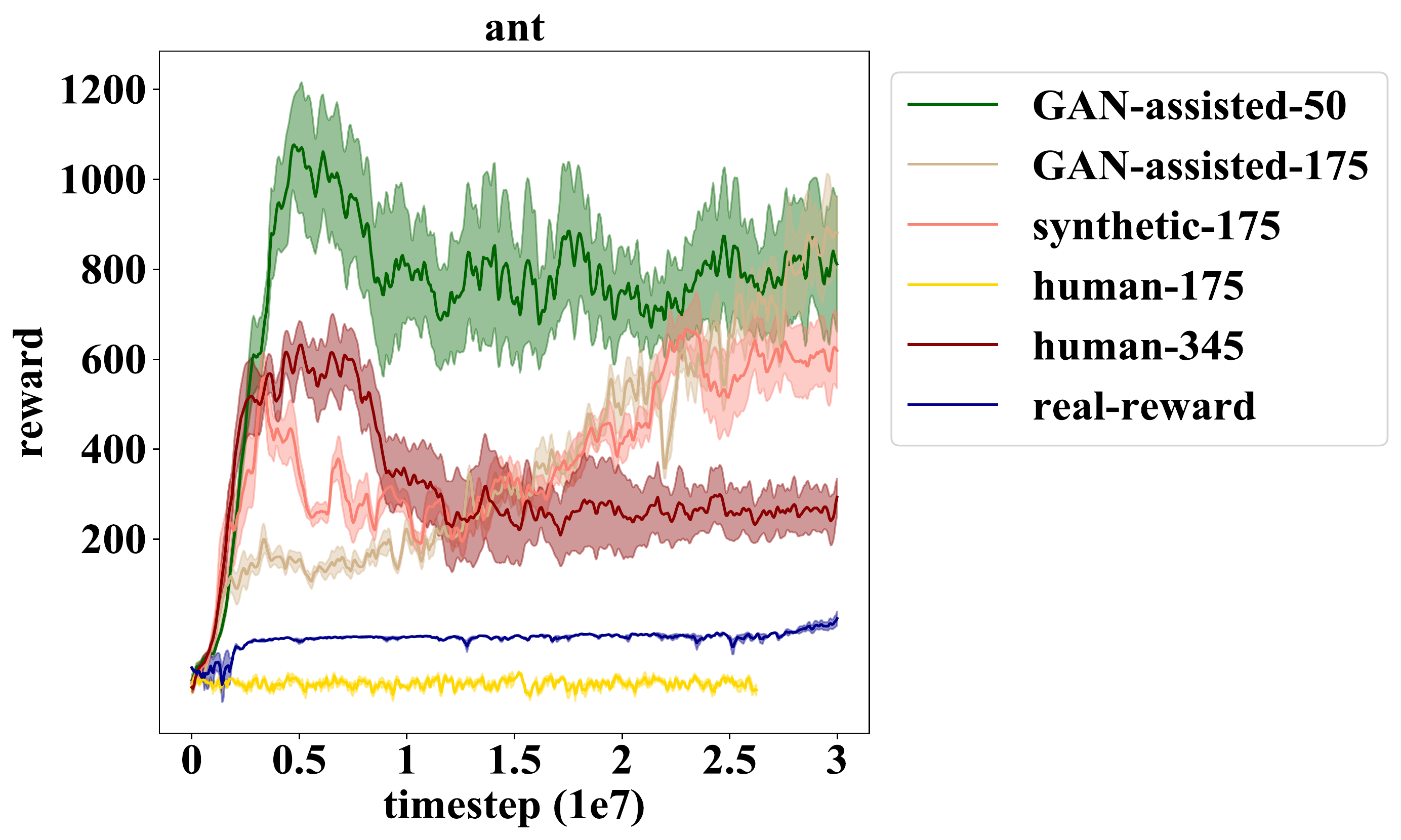}}
\caption{Comparison of numerous methods on the MuJoCo physics engine. The methods include real reward, synthetic-$175$, human-$175$, human-$345$, GAN-assisted-$50$, and GAN-assisted-$175$. All curves include six runs except for human-$175$ and human-$345$ that include three runs.}\label{fig:reward}
\end{figure*}

This subsection presents the results of the methods described earlier in this section. In particular, we conduct experiments on four types of robotic behaviors (namely, swimmer, hopper, cheetah, and ant) based on the MuJoCo physics engine. Figure~\ref{fig:reward} shows the comparison of the rewards obtained via, respectively, real reward, synthetic-$175$, human-$175$, human-$345$, GAN-assisted-$50$, and GAN-assisted-$175$. For human-$175$ and human-$345$, we conduct three runs. For all other methods, we conduct six runs. Fig.~\ref{fig:reward} shows the plots of the collective real rewards using these methods for the four types of robotic behaviors. The solid line represents the average reward while the shadow area represents the variance. Table~\ref{tab:final_re} shows the final reward comparison using these methods for the four types of robotic behaviors. It can be observed from Fig.~\ref{fig:reward} and Table~\ref{tab:final_re} that GAN-assisted-$50$ and GAN-assisted-$175$ yield much better performance than real reward, human-$175$, and human-$345$. GAN-assisted-$175$ also shows improved performance over (or very close performance as) synthetic-$175$. It can also be observed from Table~\ref{tab:final_re} that GAN-assisted-$50$ and GAN-assisted-$175$ yield similar rewards except for cheetah. Further comparison with the results in~\cite{christiano2017deep} (c.f. the last row in Table~\ref{tab:final_re}) shows that the proposed GAN-assisted-$175$ outperforms the approach proposed in~\cite{christiano2017deep} for all four cases. The proposed GAN-assisted-$50$ outperforms the approach proposed in~\cite{christiano2017deep} for three cases except for Cheetah. %The reward learning procedure assigns positive rewards to all behaviors that are typically followed by high reward.

\begin{table}[hhhh]
\caption {Performance Comparison (Final Reward)}\label{tab:final_re}
\begin{center}
\scalebox{1.1}{
\begin{tabular}{ c|c|c|c|c } 
 \hline
 Methods & Swimmer & Hopper & Cheetah & Ant\\ 
\hline
 Real Reward & 88 & 3322 & 1882 & 8 \\ 
 \hline
 Synthetic-175 & 140 & {4063}&5514&617 \\ 
 \hline
 Human-175 & 107 & 1462 &1158&-128 \\ 
Human-345 & 140 & 2758 & 2712 & 292 \\ 
\hline
GAN-assisted-50 & \textbf{167} & 3961&3485&808 \\ 
GAN-assisted-175 & 131 & \textbf{4124}& \textbf{8238}&\textbf{878} \\ 
\hline\hline
Method in~\cite{christiano2017deep} & 100 & 3900 & 5600 & -100\\
\hline
\end{tabular}}
\end{center}
\end{table}

It is also interesting to notice that human-$175$ is less effective than the typical reinforcement learning method using real reward. However, human-$345$ outperforms (or yields a similar performance as) the typical reinforcement learning method using real reward. It is our hypothesis that human provides enough correct data for reward shaping in human-$345$ while not in human-$175$ because human can make mistakes in providing preferences. Hence, the typical reinforcement learning method using real reward may outperform human-$175$ because the shaped reward may not be accurate. One unique advantage of the proposed GAN-assisted human-preference reinforcement learning approach is to separate the labeling of trajectory pairs and the creation of reinforcement learning algorithm to learn appropriate control policies. In other words, one set of human queries are used to create two sets of valuable outcomes, namely, an adversarial neural network model to directly learn human preferences and the learning of reward functions for control policy generation. Moreover, the adversarial neural network to be learned with a binary output is low-dimensional, and hence can be trained via a very small amount of human queries. In addition, GAN-based human preference queries via the learned adversarial neural network can be used to replace active human queries (required in, \textit{e.g.,}~\cite{singh2019end}) in the reward shaping and control policy generation. Hence, the proposed new GAN-assisted preference-based reinforcement learning approach can provide much more effective human preference-based reinforcement learning algorithms for the generation of trajectories that meet human preferences.   %Without the inclusion of adversarial neural network, the training of a good reward function to produce a good policy that can be used to label trajectory pairs will require more human queries. 

To demonstrate the effectiveness of the proposed GAN-assisted preference-based reinforcement learning approach, we also conducted some comparison of the human time when using the proposed approach and a baseline approach presented in~\cite{christiano2017deep}. Under the assumption that the human can respond to each query in $5$ seconds, the total human time needed in~\cite{christiano2017deep} ranges from $0.5\sim5$ hours. Correspondingly, the proposed GAN-assisted preference-based reinforcement learning approach requires only $0.08\sim0.83$ hours, further reducing about 84\% human time needed in~\cite{christiano2017deep}.

\subsection{Discussion on GAN-assisted Human Performance Learning}
\begin{figure}[t]
\centering
\vspace{-1cm}\hspace{-0.4cm}
\subfloat{\includegraphics[width=.73\columnwidth]{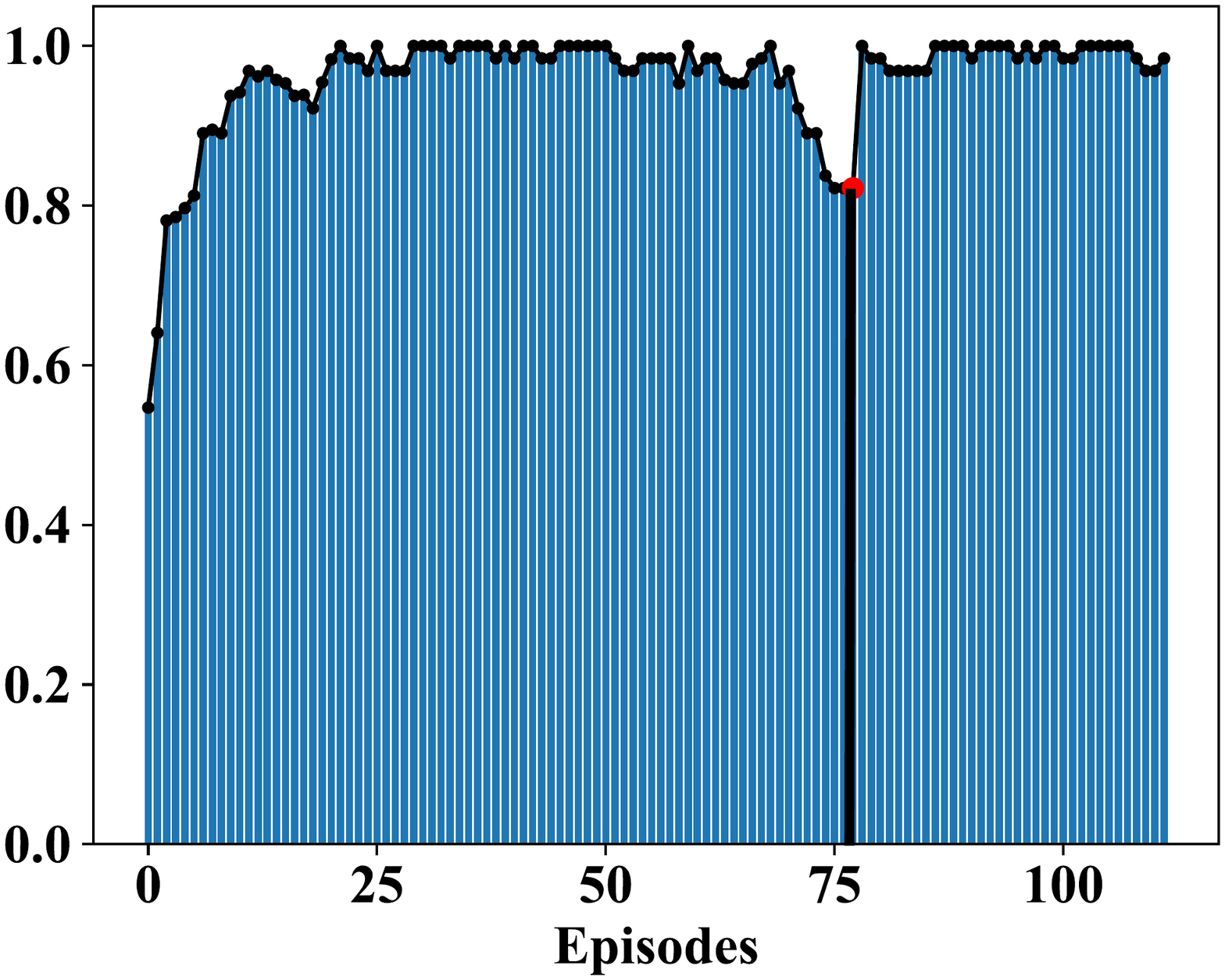}}\\ \vspace{-1.5cm}
\subfloat{\includegraphics[width=.7\columnwidth]{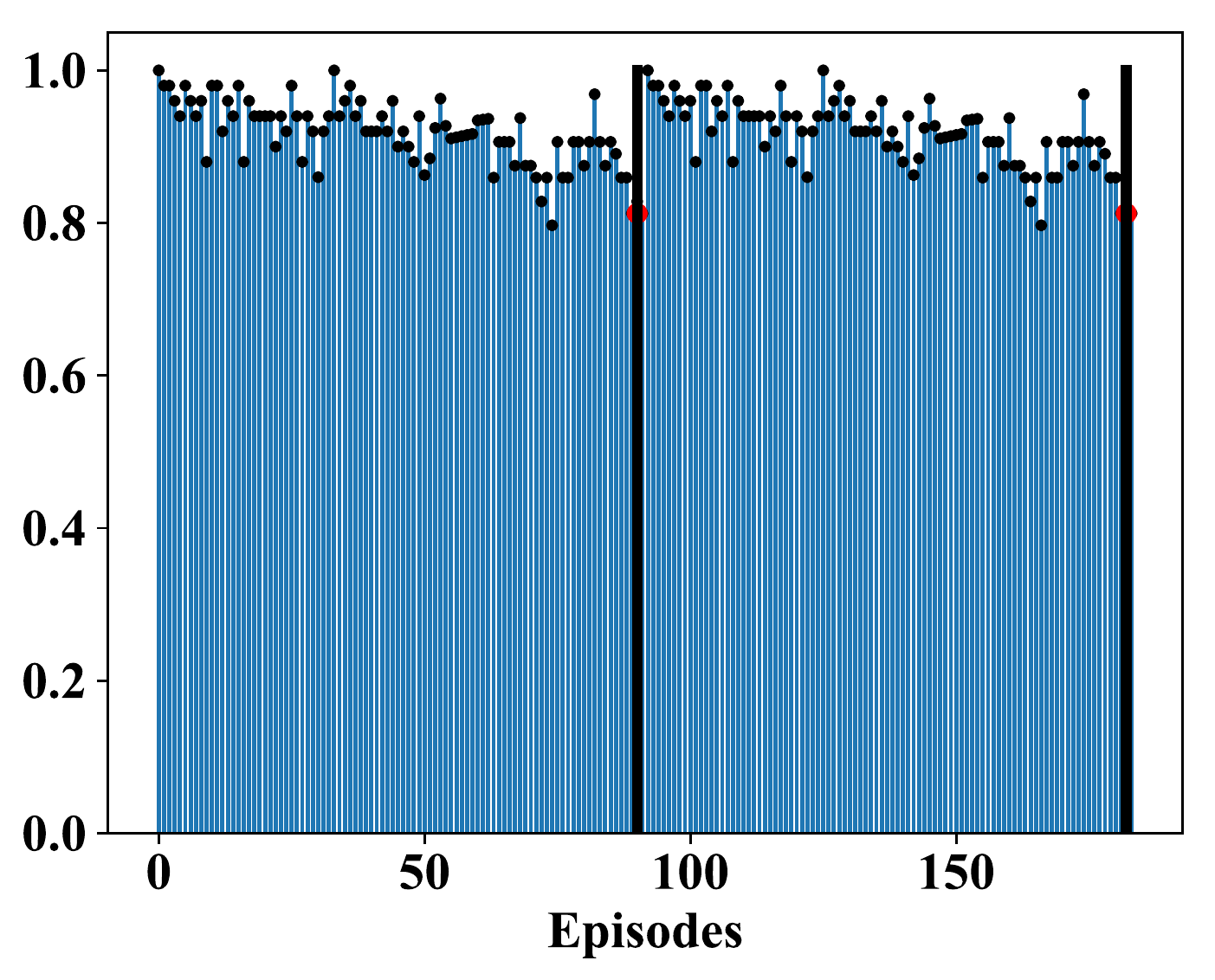}} % Reduce the figure size so that it is slightly narrower than the column. Don't use precise values for figure width.This setup will avoid overfull boxes. 
\caption{ GAN-test value for the hopper robot example (Top: GAN-assisted-175; Bottom: GAN-assisted-50). The red dots represent the \textcolor{black}{number} of episodes at which human queries \textcolor{black}{are} used.}
\label{fig:GANtEST}
\end{figure}
To quantitatively evaluate the effectiveness of the proposed GAN-assisted preference-based reinforcement learning algorithm, we now provide some discussion about its performance. In particular, we use the metric GAN-test that was developed in~\cite{shmelkov2018good}. GAN-test refers to the accuracy of a neural network, trained on physical data, evaluated on the generated data. In the context of this paper, we train the adversarial neural network on sampled trajectory pairs and then evaluate it on the newly generated trajectories to see if the output probability for human preference fits the synthetic oracle. A high GAN-test value means that the generated samples are good approximation of the (unknown) distribution of human preference-based models. An example of the GAN-test result for the hopper robot case is shown in Fig.~\ref{fig:GANtEST}. In the experiment, each episode has $4e5$ time steps. In particular, for GAN-assisted-175, we fine tune the adversarial neural network using the human preferences at episode $77$. This means that less than $1$\% human queries are used. If we assume that the human can respond to each query in $5$ seconds, a total of ${17.5}$ hours of human labeling time can be saved.  For GAN-assisted-50, we only fine tune the adversarial neural network using the human preferences at episodes $90$ and $182$. This means that only $1.1$\% human queries are used. In both cases, a high GAN-test value can be maintained, as shown in Fig.~\ref{fig:GANtEST}. One can also observe that GAN-assisted-175 yields a higher and more stable GAN-test value than GAN-assisted-50 because GAN-assisted-175 outputs a better model than GAN-assisted-50. Since GAN-assisted-175 already provides a very high GAN-test value, it is expected that more human queries may help gain very limited, if not no, performance improvements.

%\subsection{Complex Novel Behaviors}
%\begin{figure}[ht]   
%\centering
%\subfloat{\includegraphics[width=2cm]{./image/clip1.png}}
%\subfloat{\includegraphics[width=2cm]{./image/clip2.png}}
%\subfloat{\includegraphics[width=2cm]{./image/clip3.png}}
%\subfloat{\includegraphics[width=2cm]{./image/clip4.png}}\\
%\subfloat{\includegraphics[width=2cm]{./image/clip2_1.png}}
%\subfloat{\includegraphics[width=2cm]{./image/clip2_2.png}}
%\subfloat{\includegraphics[width=2cm]{./image/clip2_3.png}}
%\subfloat{\includegraphics[width=2cm]{./image/clip2_4.png}}
%\caption[Optional caption]{Demonstration of two trained novel behaviors.}\label{fig:behavior}
%\end{figure}
%Comparison of the proposed approach with the synthetic RL oracle helps us evaluate if the proposed method is effective. Another key question is whether the proposed approach can learn novel complex behaviors when no reward function is available. To answer the question, we use the same parameters in the previous experiments to evaluate if the proposed algorithm can learn novel complex behaviors by using only $50$ queries in less than {0.16} hours of human time. In particular, we demonstrate that (1) the hopper robot performs standup and jump without moving forward; \footnote{The agent learns standup and jump consistently to adjust its position to avoid falling down.} and (2) the hopper robot performs horizontal moves using its first joint.
%The two trained complex novel behaviors are shown in Figure~\ref{fig:behavior}. 

\section{Conclusion} \label{sec:con}
Effective human-robot interaction is essential for the development of sophisticated robotic systems that can act properly when interacting with real-world environments, subject to human's personalized preferences. We showed in the paper that robots can learn complex robotic behaviors with a reduction of about $99.8$\% human time (or a reduction of $84$\% human time from the baseline approach in~\cite{christiano2017deep}) when an adversarial neural network and a new maximum entropy reinforcement learning algorithm are designed appropriately. This shows the feasibility of training sophisticated robotic systems to learn human preferences efficiently. The proposed techniques and algorithms present an important step towards practical application of deep reinforcement learning in complex real-world robotic tasks when robots only need sparse human queries to learn human preferences and complex novel behaviors without access to reward functions. %The code for the proposed approach is publicly available at http://bit.ly/2TjJPko. Videos of some behaviors learned via the proposed approach can be found at http://bit.ly/2t3Dlva. 

\bibliography{Bibliography-File} 
\bibliographystyle{ieeetr}

\end{document}